\documentclass[letterpaper, 10 pt, journal, twoside]{ieeetran}

\IEEEoverridecommandlockouts
\pdfoutput=1

\usepackage{xparse} 
\usepackage{amsmath} 
\usepackage{amssymb}
\usepackage{amsfonts} 
\usepackage{empheq}

\usepackage{graphicx}
\usepackage[caption=false]{subfig}
\usepackage[keeplastbox]{flushend}

\makeatletter
\let\NAT@parse\undefined
\makeatother

\usepackage[colorlinks]{hyperref}
\usepackage{cleveref}
\usepackage{url}

\usepackage{booktabs}
\usepackage[para]{threeparttable}
\usepackage{multirow}

\usepackage[T1]{fontenc} 
\usepackage[style=ieee,backend=bibtex]{biblatex}
\AtBeginBibliography{\footnotesize}

\NewDocumentCommand\bbm{}{ \begin{bmatrix} }
\NewDocumentCommand\ebm{}{ \end{bmatrix} }
\NewDocumentCommand\Vector{m}{ \boldsymbol{\mathbf{#1}} }
\NewDocumentCommand\Matrix{m}{ \boldsymbol{\mathbf{#1}} }
\NewDocumentCommand\Transpose{m}{ \left.{#1}\right.^T }

\NewDocumentCommand\Norm{m}{\left\Vert#1\right\Vert }

\NewDocumentCommand\PartialDerivative{mm}{ \frac{\partial #1}{\partial #2} }

\NewDocumentCommand\ArgMin{m}{ \operatorname*{argmin}_{#1} }

\NewDocumentCommand\Real{}{ \mathbb{R} }

\NewDocumentCommand\LieGroupSO{m}{ \mathrm{SO}(#1) }

\NewDocumentCommand\LieGroupSE{m}{ \mathrm{SE}(#1) }

\NewDocumentCommand\NormalDistribution{mm}{ \mathcal{N}\left(#1,#2\right) }

\NewDocumentCommand\IdentityMatrix{}{ \Matrix{1} }

\NewDocumentCommand\CoordinateFrame{m}{ \underrightarrow{\Matrix{\mathcal{F}}}_{#1} }
\NewDocumentCommand\Rotation{}{ \Matrix{C} }
\NewDocumentCommand\RotationVector{}{ \Vector{\phi} }
\NewDocumentCommand\Transform{}{ \Matrix{T} }
\NewDocumentCommand\TransformVector{}{ \Vector{\xi} }

\NewDocumentCommand\Estimate{m}{\hat{#1}}

\NewDocumentCommand\Mean{m}{\overline{#1}}

\NewDocumentCommand\Image{}{\Vector{I}}

\NewDocumentCommand\deltap{}{\Delta_p}
\NewDocumentCommand\TargetVOTransform{}{\Transform_{i,i+\deltap}}
\NewDocumentCommand\EstimatedVOTransform{}{\Estimate{\Transform}_{i,i+\deltap}}
\NewDocumentCommand\EstimatedVOTransformNoSub{}{\Estimate{\Transform}}

\NewDocumentCommand\TargetCorrection{}{\Transform^*}
\NewDocumentCommand\TargetCorrectionIndex{}{\Transform^*_i}

\NewDocumentCommand\TargetCorrectionVector{}{\TransformVector^*}
\NewDocumentCommand\TargetCorrectionVectorIndex{}{\TransformVector^*_i}

\NewDocumentCommand\TargetCorrectionRotation{}{\Rotation_*}

\NewDocumentCommand\LeftJacobianSE{}{ \Matrix{\mathcal{J}} }
\NewDocumentCommand\LeftJacobianSO{}{ \Matrix{J} }
\NewDocumentCommand\LogMapFunction{m}{g({#1})}
\NewDocumentCommand\LogMapFunctionSO{m}{f({#1})}

\NewDocumentCommand\EstimatedVOTransformCorrected{}{\EstimatedVOTransform^{\text{corr}}}
\NewDocumentCommand\EstimatedVOTransformCorrectedNoSub{}{\EstimatedVOTransformNoSub^{\mathrm{corr}}}

\NewDocumentCommand\PredictionVector{}{\TransformVector}

\NewDocumentCommand\TransformCovariance{}{\Matrix{\Sigma}}

\NewDocumentCommand\CostFunction{}{\mathcal{L}}
\NewDocumentCommand\PoseCostFunction{}{\mathcal{O}}

\NewDocumentCommand\MatLn{m}{\text{ln}\left({#1} \right)^{\vee}}
\NewDocumentCommand\MatExp{m}{\text{exp}\left({#1}^{\wedge}\right)}
\NewDocumentCommand\Inv{m}{{#1}^{-1}}

\NewDocumentCommand\definedtobe{}{\triangleq}
\NewDocumentCommand\ImageQuad{}{\Image}

\newlength\fsep
\newsavebox\widebox

\NewDocumentCommand\ProjectionFunction{}{\pi}

\NewDocumentCommand\ImageCovariance{}{\Matrix{\Sigma_y}}
\NewDocumentCommand\ImageLandmark{mm}{\Vector{y}_{#1,#2}}

\title{DPC-Net: Deep Pose Correction for Visual Localization}

\author{Valentin Peretroukhin$^{1}$, and Jonathan Kelly$^{1}$%
\thanks{Manuscript received: September, 10, 2017; Revised November 24, 2017; Accepted November, 8, 2017.}
\thanks{$^{1}$Both authors are with the Space \& Terrestrial Autonomous Robotic Systems (STARS) laboratory at the University of Toronto Institute for Aerospace Studies (UTIAS), Canada. {\tt <firstname>.<lastname>@robotics.utias.utoronto.ca}}%
}

\makeatletter
\def\ps@IEEEtitlepagestyle{%
  \def\@oddfoot{\mycopyrightnotice}%
  \def\@evenfoot{\mycopyrightnotice}%
}
\def\mycopyrightnotice{%
  {\hfill 
  }
}
\makeatother

\begin{document} 

\maketitle 

\begin{abstract}
We present a novel method to fuse the power of deep networks with the computational efficiency of geometric and probabilistic localization algorithms. In contrast to other methods that completely replace a classical visual estimator with a deep network,
we propose an approach that uses a convolutional neural network to learn difficult-to-model \textit{corrections} to the estimator from ground-truth training data. To this end, we derive a novel loss function for learning $\LieGroupSE{3}$ corrections based on a matrix Lie groups approach, with a natural formulation for balancing translation and rotation errors. We use this loss to train a Deep Pose Correction network (DPC-Net) that predicts corrections for a particular estimator, sensor and environment. 
Using the KITTI odometry dataset, we demonstrate significant improvements to the accuracy of a computationally-efficient sparse stereo visual odometry pipeline, that render it as accurate as a modern computationally-intensive dense estimator. 
Further, we show how DPC-Net can be used to mitigate the effect of poorly calibrated lens distortion parameters.

\end{abstract}

\begin{IEEEkeywords}
Deep Learning in Robotics and Automation, Localization
\end{IEEEkeywords}

\markboth{IEEE Robotics and Automation Letters. Preprint Version. Accepted November, 2017}
{Peretroukhin \MakeLowercase{\textit{et al.}}: DPC-Net: Deep Pose Correction for Visual Localization} 


\section{Introduction}

\IEEEPARstart{D}{eep} convolutional neural networks (CNNs) are at the core of many state-of-the-art classification and segmentation algorithms in computer vision~\cite{LeCun2015-fw}. These CNN-based techniques achieve accuracies previously
unattainable by classical methods. In mobile robotics and state estimation, however, it remains unclear to what extent these deep architectures can obviate the need for classical geometric modelling. 
In this work, we focus on visual odometry (VO): the task of computing the egomotion of a camera through an unknown environment with no external positioning sources. Visual localization algorithms like VO can suffer from several systematic error sources
that include estimator biases \cite{Peretroukhin2014-db}, poor calibration, and environmental factors (e.g., a lack of scene texture). While machine learning approaches can be used to better model specific subsystems of a localization pipeline (e.g., the heteroscedastic feature track error covariance modelling of \cite{Peretroukhin2016-om, Peretroukhin2015-em}), much recent work \cite{Costante2016-hb, Clark2017-zg, Kendall2017-ix, Melekhov2017-dl, Oliveira2017-lt} has been devoted to completely replacing the estimator with a CNN-based system.

\begin{figure}
	\centering
	\includegraphics[width=0.48\textwidth]{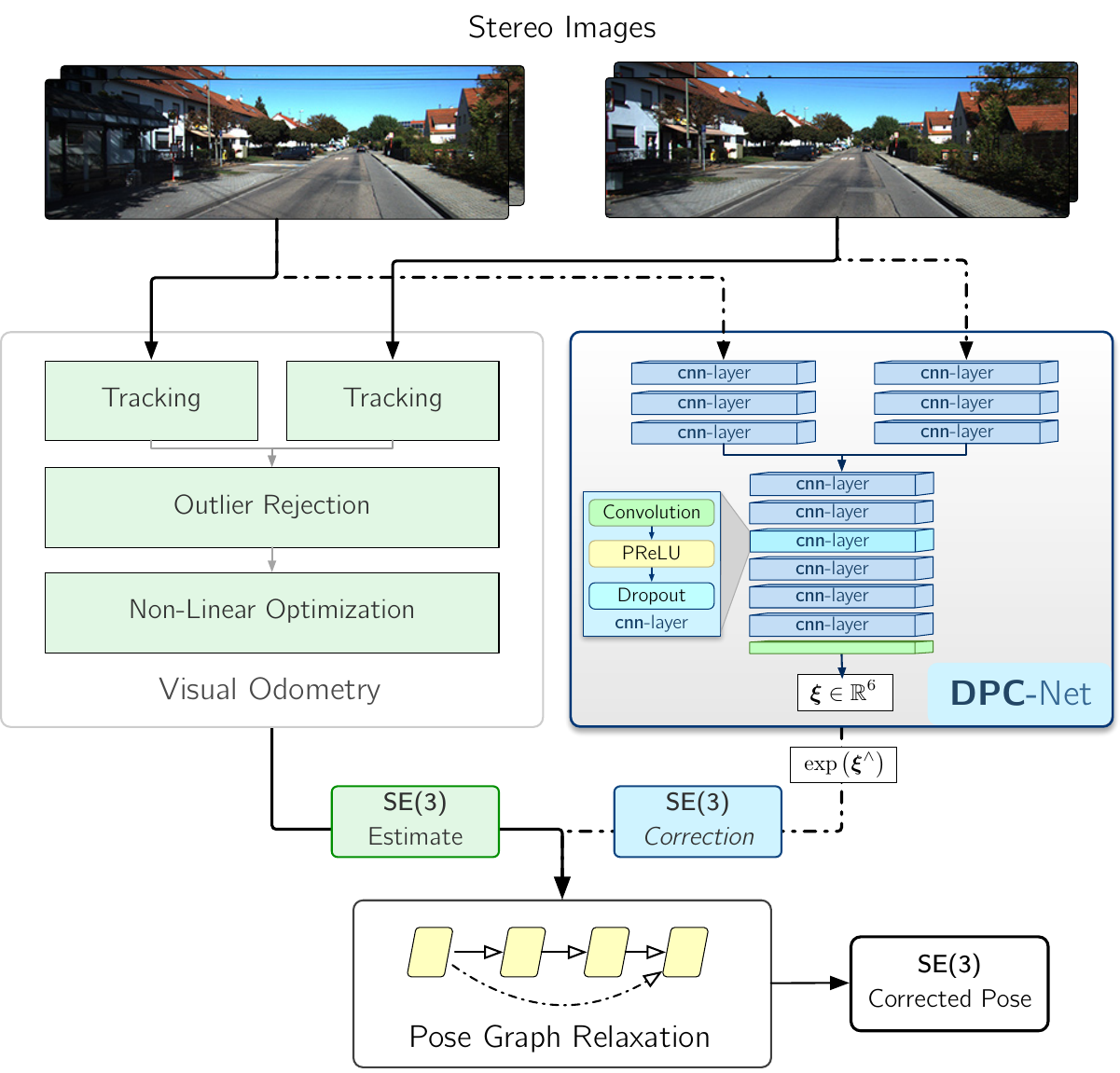}
	\caption{We propose a Deep Pose Correction network (DPC-Net) that learns $\LieGroupSE{3}$ \textit{corrections} to classical visual localizers.}
	\label{fig:system_overview}
	\vspace{-1.5em}
\end{figure}

We contend that this type of complete replacement places an unnecessary burden on the CNN. Not only must it learn projective geometry, but it must also understand the environment, and account for sensor calibration and noise.  Instead, we take inspiration from recent results \cite{Peretroukhin2017-eb} that demonstrate that CNNs can infer difficult-to-model geometric quantities (e.g., the direction of the sun) to improve an existing localization estimate.  
In a similar vein, we propose a system (\Cref{fig:system_overview}) that takes as its starting point an efficient, classical localization algorithm that computes high-rate pose estimates. To it, we add a Deep Pose Correction Network (DPC-Net) that learns low-rate, `small' \textit{corrections} from training data that we then fuse with the original estimates. DPC-Net does not require any modification to an existing localization pipeline, and can learn to correct multi-faceted errors from estimator bias, sensor mis-calibration or environmental effects.  

Although in this work we focus on visual data, the DPC-Net architecture can be readily modified to learn $\LieGroupSE{3}$ corrections for estimators that operate with other sensor modalities (e.g., lidar). For this general task, we derive a pose regression loss function and a closed-form analytic expression for its Jacobian. Our loss permits a network to learn an unconstrained Lie algebra coordinate vector, but derives its Jacobian with respect to $\LieGroupSE{3}$ geodesic distance. 

In summary, the main contributions of this work are
\begin{enumerate}
	\item the formulation of a novel deep corrective approach to egomotion estimation,
	\item a novel cost function for deep $\LieGroupSE{3}$ regression that naturally balances translation and rotation errors, and
	\item an open-source implementation of DPC-Net in \texttt{PyTorch}\footnote{See \url{https://github.com/utiasSTARS/dpc-net}.}.
	
\end{enumerate}

\section{Related Work}
Visual state estimation has a rich history in mobile robotics. We direct the reader to \cite{Scaramuzza2011-qr} and \cite{Cadena2016-ds} for detailed surveys of what we call \textit{classical}, geometric approaches to visual odometry (VO) and visual simultaneous localization and mapping (SLAM).

In the past decade, much of machine learning and its sub-disciplines has been revolutionized by carefully constructed deep neural networks (DNNs)~\cite{LeCun2015-fw}. For tasks like image segmentation, classification, and natural language processing, most prior state-of-the-art algorithms have been replaced by their DNN alternatives.

In mobile robotics, deep neural networks have ushered in a new paradigm of end-to-end training of visuomotor policies \cite{Levine2016-bj} and significantly impacted the related field of reinforcement learning~\cite{Duan2016-jw}. In state estimation, however, most successful applications of deep networks have aimed at replacing a specific sub-system of a localization and mapping pipeline (for example, object detection~\cite{yang2016exploit}, place recognition~\cite{Sunderhauf2015-oc}, or bespoke discriminative observation functions~\cite{Haarnoja2016-ph}).

Nevertheless, a number of recent approaches have presented convolutional neural network architectures that purport to obviate the need for classical visual localization algorithms. For example, Kendall et al.~\cite{Kendall2015-kh,Kendall2017-ix} presented extensive work on PoseNet, a CNN-based camera re-localization approach that regresses the 6-DOF pose of a camera within a previously explored environment. Building upon PoseNet, Melekhov~\cite{Melekhov2017-dl} applied a similar CNN learning paradigm to relative camera motion. In related work, Costante et al. \cite{Costante2016-hb} presented a CNN-based VO technique that uses pre-processed dense optical flow images. By focusing on RGB-D odometry, Handa et al.~\cite{Handa2016-hm}, detailed an approach to learning relative poses with \textit{3D Spatial Transformer} modules and a dense photometric loss. Finally, Oliviera et al. \cite{Oliveira2017-lt} and Clark et al.~\cite{Clark2017-zg} described techniques for more general sensor fusion and mapping. The former work outlined a DNN-based topometric localization pipeline with separate VO and place recognition modules while the latter presented VINet, a CNN paired with a recurrent neural network for visual-inertial sensor fusion and online calibration.

With this recent surge of work in end-to-end learning for visual localization, one may be tempted to think this is the only way forward. It is important to note, however, that these deep CNN-based approaches do not yet report state-of-the-art localization accuracy, focusing instead on proof-of-concept validation. Indeed, at the time of writing, the most accurate visual odometry approach on the KITTI odometry benchmark leaderboard\footnote{See \url{http://www.cvlibs.net/datasets/kitti/eval_odometry.php}.} remains a variant of a sparse feature-based pipeline with carefully hand-tuned optimization~\cite{Cvisic2015-mt}. 

Taking inspiration from recent results that show that CNNs can be used to inject global orientation information into a visual localization pipeline \cite{Peretroukhin2017-eb}, and leveraging ideas from recent approaches to trajectory tracking in the field of controls~\cite{2017_Li_Deep, Punjani2015-pj}, we formulate a system that learns \textit{pose corrections} to an existing estimator, instead of learning the entire localization problem \textit{ab initio}.

\begin{figure}
	\centering
	\includegraphics[width=0.48\textwidth]{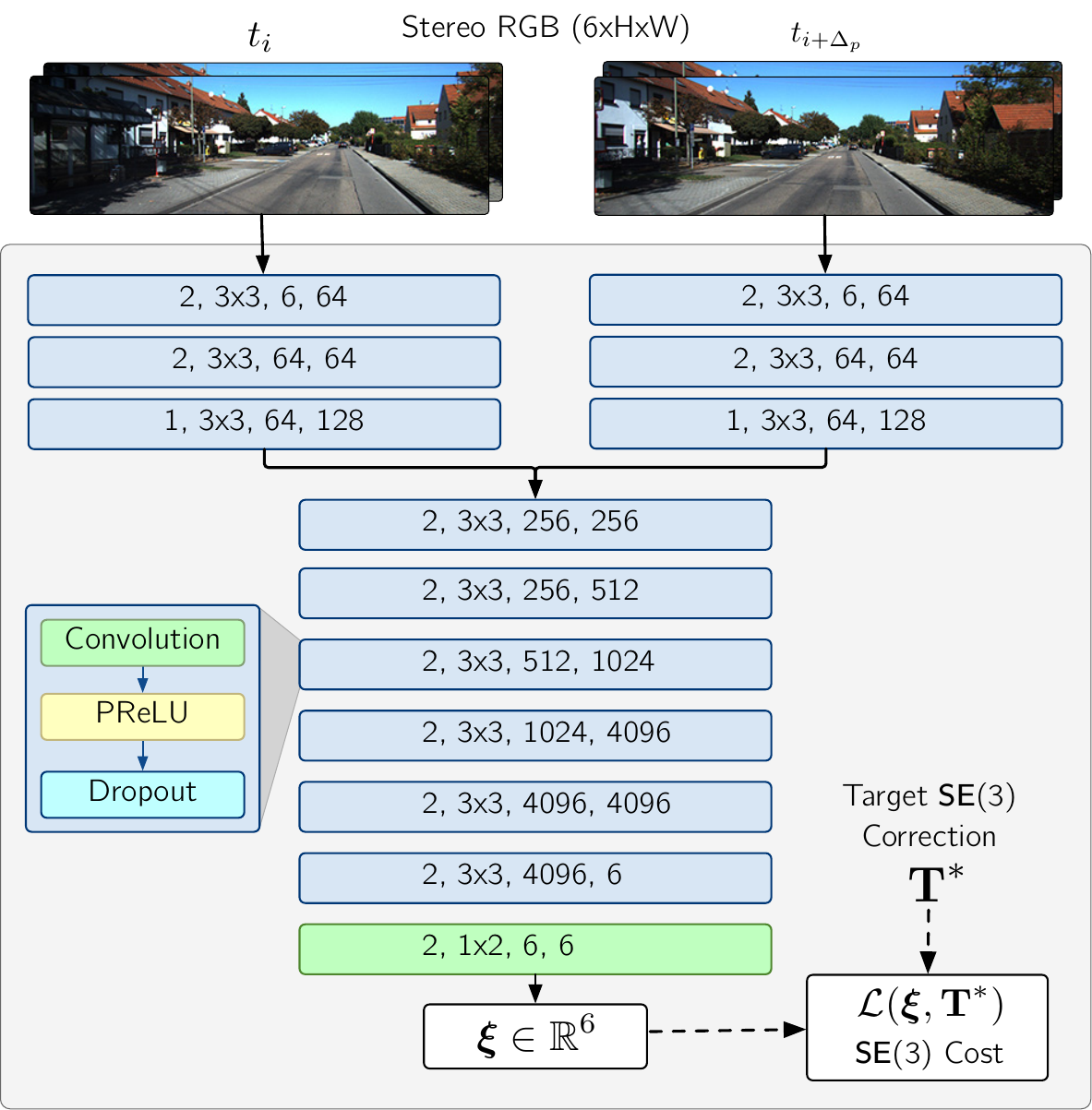}
	\caption{The Deep Pose Correction network with stereo RGB image data. 
	The network learns a map from two stereo pairs to a vector of Lie algebra coordinates. 
	Each darker blue block consists of a convolution, a PReLU non-linearity, and a dropout layer. We opt to not use MaxPooling in the network, following \cite{Handa2016-hm}.
	The labels correspond to the stride, kernel size, and number of input and output channels of for each convolution.}
	\label{fig:dpc_net}
	\vspace{-1em}
\end{figure}

\section{System Overview: Deep Pose Correction}

We base our network structure on that of \cite{Handa2016-hm} but learn $\LieGroupSE{3}$ \textit{corrections} from stereo images and require no pre-training. Similar to \cite{Handa2016-hm}, we use primarily \texttt{3x3} convolutions and avoid the use of max pooling to preserve spatial information. We achieve downsampling by setting the stride to 2 where appropriate (see \Cref{fig:dpc_net} for a full description of the network structure). We derive a novel loss function, unlike that used in \cite{Kendall2017-ix, Melekhov2017-dl, Oliveira2017-lt}, based on $\LieGroupSE{3}$ geodesic distance. Our loss naturally balances translation and rotation error without requiring a hand-tuned scalar hyper-parameter. Similar to \cite{Costante2016-hb}, we test our final system on the KITTI odometry benchmark, and evaluate how it copes with degraded visual data. 

Given two coordinate frames $\CoordinateFrame{i}$, $\CoordinateFrame{i+\deltap}$ that represent a camera's pose at time $t_{i}$ and $t_{i+\deltap}$ (where $\deltap$ is an integer hyper-parameter that allows DPC-Net to learn corrections across multiple temporally consecutive poses), we assume that our visual localizer gives us an estimate\footnote{We use $\Estimate{(\cdot)}$ to denote an estimated quantity throughout the paper.}, $\EstimatedVOTransform$, of the true transform $\TargetVOTransform \in \LieGroupSE{3}$ between the two frames. We aim to learn a target correction,
\begin{equation}
	\TargetCorrectionIndex = \TargetVOTransform \Inv{\EstimatedVOTransform},
\end{equation}
from two pairs of stereo images (collectively referred to as $\ImageQuad_{t_{i}, t_{i+\deltap}}$) captured at $t_{i}$ and $t_{i+\deltap}$\footnote{Note that the visual estimator does not necessarily compute $\EstimatedVOTransform$ directly. $\EstimatedVOTransform$ may be compounded from several estimates.}. Given a dataset, $\{ \TargetCorrectionIndex, \ImageQuad_{t_{i}, t_{i+\deltap}} \}^N_{i=1}$,  we now turn to the problem of selecting an appropriate loss function for learning $\LieGroupSE{3}$ corrections.

\subsection{Loss Function: Correcting $\LieGroupSE{3}$ Estimates}

One possible approach to learning $\TargetCorrection$ is to break it into constituent parts and then compose a loss function as the weighted sum of translational and rotational error norms (as done in \cite{Kendall2017-ix, Melekhov2017-dl, Oliveira2017-lt}). This however does not account for the possible correlation between the two losses, and requires the careful tuning of a scalar weight.

In this work, we instead choose to parametrize our correction prediction as $\Transform = \MatExp{\PredictionVector}$, where $\PredictionVector \in \Real^6$, a vector of Lie algebra coordinates, is the output of our network (similar to \cite{Handa2016-hm}). We define a loss for $\PredictionVector$ as 
\begin{equation}
	\label{eq:se3loss_fn}
\CostFunction (\PredictionVector) = \frac{1}{2} \LogMapFunction{\PredictionVector}^T \Inv{\TransformCovariance} \LogMapFunction{\PredictionVector},
\end{equation} where 
\begin{equation}
	\label{eq:se3logmap_fn}
	\LogMapFunction{\PredictionVector} \definedtobe \MatLn{\MatExp{\PredictionVector} \Inv{\TargetCorrection}}.
\end{equation}
Here, $\TransformCovariance$ is the covariance of our estimator (expressed using unconstrained Lie algebra coordinates), and $(\cdot)^\wedge$, $(\cdot)^\vee$ convert vectors of Lie algebra coordinates to matrix vectorspace quantities and back, respectively.
Given two stereo image pairs, we use the output of DPC-Net, $\PredictionVector$, to correct our estimator as follows:
\begin{align}
	\EstimatedVOTransformCorrectedNoSub = \MatExp{\PredictionVector} \EstimatedVOTransformNoSub,
\end{align}
where we have dropped subscripts for clarity.

\subsection{Loss Function: $\LieGroupSE{3}$ Covariance}

Since we are learning estimator \textit{corrections}, we can compute an empirical covariance over the training set as

\begin{equation}
\TransformCovariance = \frac{1}{N-1} \sum_{i=1}^{N} \left( \TargetCorrectionVectorIndex -\Mean{\TargetCorrectionVector} \right) \left( \TargetCorrectionVectorIndex -\Mean{\TargetCorrectionVector} \right)^T,
\end{equation}
where 
\begin{equation}
	\TargetCorrectionVectorIndex \definedtobe \MatLn{\TargetCorrectionIndex}, \quad \Mean{\TargetCorrectionVector} \definedtobe \frac{1}{N} \sum_{i=1}^N \TargetCorrectionVectorIndex.
\end{equation}

The term $\TransformCovariance$ balances the rotational and translation loss terms based on their magnitudes in the training set, and accounts for potential correlations. We stress that if we were learning poses directly, the pose targets and their associated mean would be trajectory dependent and would render this type of covariance estimation meaningless. Further, we find that, in our experiments, $\TransformCovariance$ weights translational and rotational errors similarly to that presented in \cite{Kendall2017-ix} 
 based on the diagonal components, but contains relatively large off-diagonal terms.

\subsection{Loss Function: $\LieGroupSE{3}$ Jacobians}
In order to use \Cref{eq:se3loss_fn} to train DPC-Net with back-propagation, we need to compute its Jacobian with respect to our network output, $\PredictionVector$. Applying the chain rule, we begin with the expression
\begin{equation}
	\label{eq:se3_jacobchainrule}
	\PartialDerivative{\CostFunction (\PredictionVector) }{ \PredictionVector} =  \LogMapFunction{\PredictionVector}^T \Inv{\TransformCovariance} \PartialDerivative{\LogMapFunction{\PredictionVector}}{\PredictionVector}.
\end{equation}
The term $\PartialDerivative{\LogMapFunction{\PredictionVector}}{\PredictionVector}$ is of importance. We can derive it in two ways. To start, note two important identities \cite{Barfoot2017-ri}. First,
\begin{equation}
	\label{eq:delta_xi_exp}
\MatExp{( \TransformVector + \delta \TransformVector)} \approx \MatExp{(\LeftJacobianSE \delta \TransformVector)} \MatExp{\TransformVector},
\end{equation}
where $\LeftJacobianSE \definedtobe \LeftJacobianSE(\TransformVector)$ is the left $\LieGroupSE{3}$ Jacobian.
Second, if $\Transform_1 \definedtobe \MatExp{\TransformVector_1}$ and $\Transform_2 \definedtobe \MatExp{\TransformVector_2}$, then
\begin{align}
	\label{eq:jacobian_xi_small}
\MatLn{\Transform_1 \Transform_2} &= \MatLn{ \MatExp{\TransformVector_1} \MatExp{\TransformVector_2} } \nonumber \\
								  &\approx \left\{
	\begin{array}{ll}
		\Inv{\LeftJacobianSE(\TransformVector_2)} \TransformVector_1 + \TransformVector_2   & \mbox{if } \PredictionVector_1 ~ \mbox{small} \\
		\TransformVector_1 + \Inv{\LeftJacobianSE(-\TransformVector_1)} \TransformVector_2 & \mbox{if } \PredictionVector_2 ~ \mbox{small}.
	\end{array}
\right\}
\end{align}
See \cite{Barfoot2017-ri} for a detailed treatment of matrix Lie groups and their use in state estimation.
\subsubsection{Deriving $\PartialDerivative{\LogMapFunction{\PredictionVector}}{\PredictionVector}$, Method I}
If we assume that only $\TransformVector$ is `small', we can apply \Cref{eq:jacobian_xi_small} directly to define
\begin{equation}
	\label{eq:logjacob_method1}
	\PartialDerivative{\LogMapFunction{\PredictionVector}}{\PredictionVector} = \Inv{\LeftJacobianSE(-\TargetCorrectionVector)}, 
\end{equation}
with $	\TargetCorrectionVector \definedtobe \MatLn{\TargetCorrection}$. Although attractively compact, note that this expression for $\PartialDerivative{\LogMapFunction{\PredictionVector}}{\PredictionVector}$ assumes that $\PredictionVector$ is small, and may be inaccurate for `larger' $\TargetCorrection$ (since we will therefore require $\PredictionVector$ to be commensurately `large').

\subsubsection{Deriving $\PartialDerivative{\LogMapFunction{\PredictionVector}}{\PredictionVector}$, Method II}

Alternatively, we can linearize \Cref{eq:se3logmap_fn} about $\PredictionVector$, by considering a small change $\delta \PredictionVector$ and applying \Cref{eq:delta_xi_exp}:
\begin{align}
	\LogMapFunction{\PredictionVector + \delta \PredictionVector} &=  \MatLn{\MatExp{( \TransformVector + \delta \TransformVector)} \Inv{\TargetCorrection}} \\
	&\approx \MatLn{\MatExp{(\LeftJacobianSE \delta \TransformVector)} \MatExp{\TransformVector} \Inv{\TargetCorrection}}.
\end{align}

\noindent Now, assuming that $\LeftJacobianSE \delta \TransformVector$ is `small', and using \Cref{eq:se3logmap_fn}, \Cref{eq:jacobian_xi_small} gives: 
\begin{equation}
	\LogMapFunction{\PredictionVector + \delta \PredictionVector} \approx \Inv{\LeftJacobianSE(\LogMapFunction{\PredictionVector})} \LeftJacobianSE( \TransformVector) \delta \TransformVector + \LogMapFunction{\PredictionVector}.
\end{equation}
Comparing this to the first order Taylor expansion: $\LogMapFunction{\PredictionVector + \delta \PredictionVector} \approx \LogMapFunction{\PredictionVector} +  \PartialDerivative{\LogMapFunction{\PredictionVector}}{\PredictionVector} \delta \TransformVector$, we see that
\begin{equation}
	\label{eq:logjacob_method2}
	\PartialDerivative{\LogMapFunction{\PredictionVector}}{\PredictionVector} = \Inv{\LeftJacobianSE(\LogMapFunction{\PredictionVector})} \LeftJacobianSE( \TransformVector).
\end{equation}
Although slightly more computationally expensive, this expression makes no assumptions about the `magnitude` of our correction and works reliably for any target. Note further that if $\PredictionVector$ is small, then $\LeftJacobianSE( \TransformVector) \approx \IdentityMatrix$ and $\MatExp{\TransformVector} \approx \IdentityMatrix$. Thus,
\begin{equation}
	\LogMapFunction{\PredictionVector} \approx \MatLn{\Inv{\TargetCorrection}} = -\TargetCorrectionVector,
\end{equation}
and \Cref{eq:logjacob_method2} becomes
\begin{equation}
	\PartialDerivative{\LogMapFunction{\PredictionVector}}{\PredictionVector} = \Inv{\LeftJacobianSE(-\TargetCorrectionVector)}, 
\end{equation}
which matches \textit{Method I}. To summarize, to apply back-propagation to \Cref{eq:se3loss_fn}, we use \Cref{eq:se3_jacobchainrule} and \Cref{eq:logjacob_method2}.


\subsection{Loss Function: Correcting $\LieGroupSO{3}$ Estimates}
Our framework can be easily modified to learn $\LieGroupSO{3}$ corrections only. We can parametrize a similar objective for $\RotationVector \in \Real^3$,
\begin{equation}
	\label{eq:so3loss_fn}
\CostFunction (\RotationVector, \TargetCorrectionRotation) = \frac{1}{2} \LogMapFunctionSO{\RotationVector}^T \Inv{\TransformCovariance} \LogMapFunctionSO{\RotationVector},
\end{equation}
where 
\begin{equation}
	\label{eq:s03logmap_fn}
	\LogMapFunctionSO{\RotationVector} \definedtobe \MatLn{\MatExp{\RotationVector} \Inv{\TargetCorrectionRotation}}.
\end{equation}
\Cref{eq:delta_xi_exp,eq:jacobian_xi_small} have analogous $\LieGroupSO{3}$ formulations:

\begin{equation}
\label{eq:delta_phi_exp}
\MatExp{( \RotationVector + \delta \RotationVector)} \approx \MatExp{(\LeftJacobianSO \delta \RotationVector)} \MatExp{\RotationVector},
\end{equation}
and
\begin{align}
\label{eq:jacobian_phi_small}
\MatLn{\Rotation_1 \Rotation_2} &= \MatLn{ \MatExp{\RotationVector_1} \MatExp{\RotationVector_2} } \nonumber \\
								  &\approx \left\{
	\begin{array}{ll}
		\Inv{\LeftJacobianSO(\RotationVector_2)} \RotationVector_1 + \RotationVector_2   & \mbox{if } \RotationVector_1 ~ \mbox{small} \\
		\RotationVector_1 + \Inv{\LeftJacobianSO(-\RotationVector_1)} \RotationVector_2 & \mbox{if } \RotationVector_2 ~ \mbox{small},
	\end{array}
\right.
\end{align}
where $\LeftJacobianSO \definedtobe \LeftJacobianSO(\RotationVector)$ is the left $\LieGroupSO{3}$ Jacobian. Accordingly, the final loss Jacobians are identical in structure to \Cref{eq:se3_jacobchainrule} and \Cref{eq:logjacob_method2}, with the necessary $\LieGroupSO{3}$ replacements.

\subsection{Pose Graph Relaxation}

\begin{figure}
	\centering
	\includegraphics[width=0.25\textwidth]{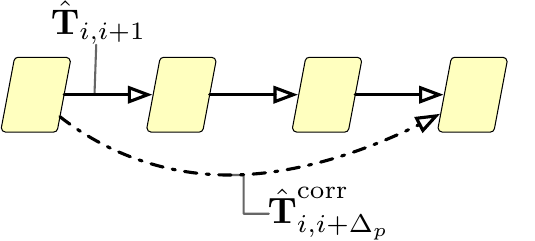}
	\caption{We apply pose graph relaxation to fuse high-rate visual localization estimates ($\Estimate{\Transform}_{i,i+1}$) with low-rate deep pose corrections ($\EstimatedVOTransformCorrected$).}
	\label{fig:posegraph_relaxation}
	\vspace{-1em}
\end{figure}

In practice, we find that using camera poses several frames apart (i.e. $\deltap > 1$) often improves test accuracy and reduces overfitting. As a result, we turn to pose graph relaxation to fuse low-rate corrections with higher-rate visual pose estimates. For a particular window of $\deltap + 1$ poses (see \Cref{fig:posegraph_relaxation}), we solve the non linear minimization problem
\begin{equation}
	\{\Transform_{t_i, n} \}_{i=0}^{\deltap} = \ArgMin{\{\Transform_{t_i, n} \}_{i=0}^{\deltap} \in \LieGroupSE{3}} \PoseCostFunction_t,
\end{equation}
where $n$ refers to a common navigation frame, and where we define the total cost, $\PoseCostFunction_t$, as a sum of visual estimation and correction components:
\begin{equation}
	\PoseCostFunction_t = \PoseCostFunction_v + \PoseCostFunction_c.
\end{equation}

\noindent The former cost sums over each estimated transform,
\begin{equation}
	\PoseCostFunction_v \definedtobe \sum_{i=0}^{\deltap-1}\Transpose{\Vector{e}}_{t_i,t_{i+1}}\TransformCovariance_v^{-1} \Vector{e}_{t_i,t_{i+1}},
\end{equation}
while the latter incorporates a single pose correction,
\begin{equation}
	\PoseCostFunction_c \definedtobe \Transpose{\Vector{e}}_{t_0,t_{\deltap}}\TransformCovariance_c^{-1} \Vector{e}_{t_0,t_{\deltap}},
\end{equation}
with the pose error defined as 
\begin{equation}
	\Vector{e}_{1,2} = \MatLn{\Estimate{\Transform}_{1,2} \Transform_{2}  \Transform^{-1}_{1}}.
\end{equation}
We refer the reader to \cite{Barfoot2017-ri} for a detailed treatment of pose-graph relaxation.

\section{Experiments}
To assess the power of our proposed deep corrective paradigm, we trained DPC-Net on visual data with localization estimates from a sparse stereo visual odometry (S-VO) estimator. In addition to training the full $\LieGroupSE{3}$ DPC-Net, we modified the loss function and input data to learn simpler $\LieGroupSO{3}$ rotation corrections, and simpler still, yaw angle corrections. For reference, we compared S-VO with different DPC-Net corrections to a state-of-the-art dense estimator. Finally, we trained DPC-Net on visual data and localization estimates from radially-distorted and cropped images. 

\subsection{Training \& Testing}
For all experiments, we used the KITTI odometry benchmark training set \cite{Geiger2013-ky}. Specifically, our data consisted of the eight sequences \texttt{00},\texttt{02} and \texttt{05}-\texttt{10} (we removed sequences \texttt{01}, \texttt{03}, \texttt{04} to ensure that all data originated from the `residential' category for training and test consistency).

For testing and validation, we selected the first three sequences (\texttt{00},\texttt{02}, and \texttt{05}). For each test sequence, we selected the subsequent sequence for validation (i.e., for test sequence \texttt{00} we validated with \texttt{02}, for \texttt{02} with \texttt{05}, etc.) and used the remaining sequences for training.  We note that by design, we train DPC-Net to predict corrections for a specific sensor and estimator pair. A pre-trained DPC-Net may further serve as a useful starting point to fine-tune new models for other sensors and estimators. In this work, however, we focus on the aforementioned KITTI sequences and leave a thorough investigation of generalization for future work.

To collect training samples, $\{ \TargetCorrectionIndex, \ImageQuad_{t_{i}, t_{i+\deltap}} \}^N_{i=0}$, we used a stereo visual odometry estimator and GPS-INS ground-truth from the KITTI odometry dataset\footnote{We used RGB stereo images to train DPC-Net but grayscale images for the estimator.}. We resized all images to ${[400, 120]}$ pixels, approximately preserving their original aspect ratio\footnote{Because our network is fully convolutional, it can, in principle, operate on different image resolutions with no modifications, though we do not investigate this ability in this work.}. For non-distorted data, we use $\deltap \in {[3,4,5]}$ for training, and test with $\deltap = 4$. For distorted data, we reduce this to $\deltap \in {[2,3,4]}$ and $\deltap = 3$, respectively, to compensate for the larger estimation errors.

Our training datasets contained between 35,000 and 52,000 training samples\footnote{If a sequence has $M$ poses, we collect $M - \deltap$ training samples for each $\deltap$.} depending on test sequence. We trained all models for 30 epochs using the Adam optimizer, and selected the best epoch based on the lowest validation loss.

\subsubsection{Rotation}
To train rotation-only corrections, we extracted the $\LieGroupSO{3}$ component of  $\TargetCorrectionIndex$ and trained our network using \Cref{eq:so3loss_fn}. Further, owing to the fact that rotation information can be extracted from monocular images, we replaced the input stereo pairs in DPC-Net with monocular images from the left camera\footnote{The \textit{stereo} VO estimator remained unchanged.}.

\subsubsection{Yaw}
To further simplify the corrections, we extracted a single-degree-of-freedom yaw rotation correction angle\footnote{We define yaw in the camera frame as the rotation about the camera's vertical $y$ axis.} from $\TargetCorrectionIndex$, and trained DPC-Net with monocular images and a mean squared loss.


\subsection{Estimators}
\subsubsection{Sparse Visual Odometry}
To collect $ \TargetCorrectionIndex$, we first used a frame-to-frame sparse visual odometry pipeline similar to that presented in \cite{Peretroukhin2016-om}. We briefly outline the pipeline here. 

Using the open-source \texttt{libviso2} package~\cite{Geiger2011-xe}, we detect and track sparse stereo image key-points, $\ImageLandmark{l}{t_{i+1}}$ and $\ImageLandmark{l}{t_{i}}$, between stereo image pairs (assumed to be undistorted and rectified).
We model reprojection errors (due to sensor noise
and quantization) as zero-mean Gaussians with a known covariance, $\ImageCovariance$,
\begin{align}
 \Vector{e}_{l,t_i} &= \ImageLandmark{l}{t_{i+1}} - \ProjectionFunction( \Transform_{t_{i+1},t_i} 
    \ProjectionFunction^{-1}( \ImageLandmark{l}{t_{i}} ) ) \\ &\sim \NormalDistribution{\Vector{0}}{\ImageCovariance},
   \label{eq:image_error}
\end{align}
where $\ProjectionFunction(\cdot)$ is the stereo camera projection function. To generate an initial guess and to reject outliers, we use three point Random Sample Consensus (RANSAC) based on stereo reprojection error.
Finally, we solve for the maximum likelihood transform, $\Transform_{t+1,t}^*$, through a Gauss-Newton minimization:
\begin{equation}
  \Transform_{t_{i+1},t_i}^* = \ArgMin{\Transform_{t_{i+1},t_i}\in\text{SE}(3)}\sum_{l=1}^{N_{t_i}} 
  \Transpose{\Vector{e}}_{l,t_i} \ImageCovariance^{-1} \Vector{e}_{l,t_i}.
\end{equation}

\subsubsection{Sparse Visual Odometry with Radial Distortion}

\begin{figure*}
	\centering
	\includegraphics[width=0.96\textwidth]{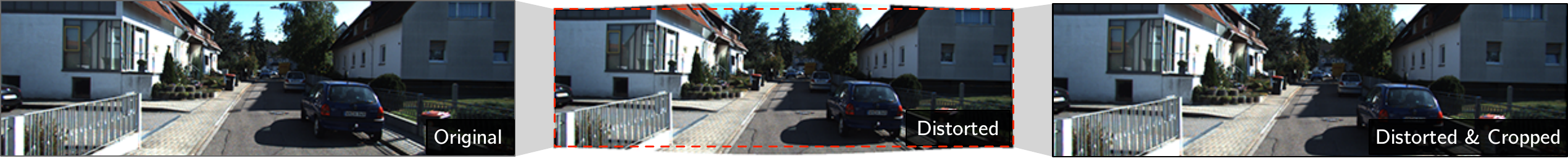}
	\caption{Illustration of our image radial distortion procedure. Left: rectified RGB image  (frame \texttt{280} from KITTI odometry sequence \texttt{05}). Middle: the same image with radial distortion applied. Right: distorted, cropped, and scaled image.}
	\label{fig:distorted_images}
	\vspace{-0.4cm}
\end{figure*}

Similar to \cite{Costante2016-hb}, we modified our input images to test our network's ability to correct estimators that compute poses from degraded visual data. Unlike \cite{Costante2016-hb}, who darken and blur their images, we chose to simulate a poorly calibrated lens model by applying radial distortion to the (rectified) KITTI dataset using a plumb-bob distortion model. The model computes radially-distorted image coordinates, $x_d, y_d$, from the normalized coordinates $x_n, y_n$ as
\begin{equation}
	\label{eq:radial_distortion}
	\bbm x_d \\ y_d \ebm = \left( 1 + \kappa_1 r^2 + \kappa_2 r^4 + \kappa_3 r^6  \right) \bbm x_n \\ y_n \ebm,
\end{equation}
where $r = \sqrt{x_n^2 + y_n^2}$. We set the distortion coefficients, $\kappa_1$, $\kappa_2$, and $\kappa_3$ to $-0.3,0.2,0.01$ respectively, to approximately match the KITTI radial distortion parameters. We solved \Cref{eq:radial_distortion} iteratively and used bilinear interpolation to compute the distorted images for every stereo pair in a sequence. Finally, we cropped each distorted image to remove any whitespace. \Cref{fig:distorted_images} illustrates this process. 

With this distorted dataset, we computed S-VO localization estimates and then trained DPC-Net to correct for the effects of the radial distortion and effective intrinsic parameter shift due to the cropping process.

\subsubsection{Dense Visual Odometry}
Finally, we present localization estimates from a computationally-intensive keyframe-based dense, direct visual localization pipeline~\cite{Clement2018-cat} that computes relative camera poses by minimizing photometric error with respect to a keyframe image. To compute the photometric error, the pipeline relies on an inverse compositional approach to map the image coordinates of a tracking image to the image coordinates of the reference depth image. As the camera moves through an environment, a new keyframe depth image is computed and stored when the camera field-of-view differs sufficiently from the last keyframe. 

We used this dense, direct estimator as our benchmark for a state-of-the-art visual localizer, and compared its accuracy to that of a much less computationally expensive sparse estimator paired with DPC-Net.

\begin{figure*}[h!]
	\centering
  	\subfloat[Sequence \texttt{00}: Cumulative Errors.]{
	   \includegraphics[width=0.5\linewidth]{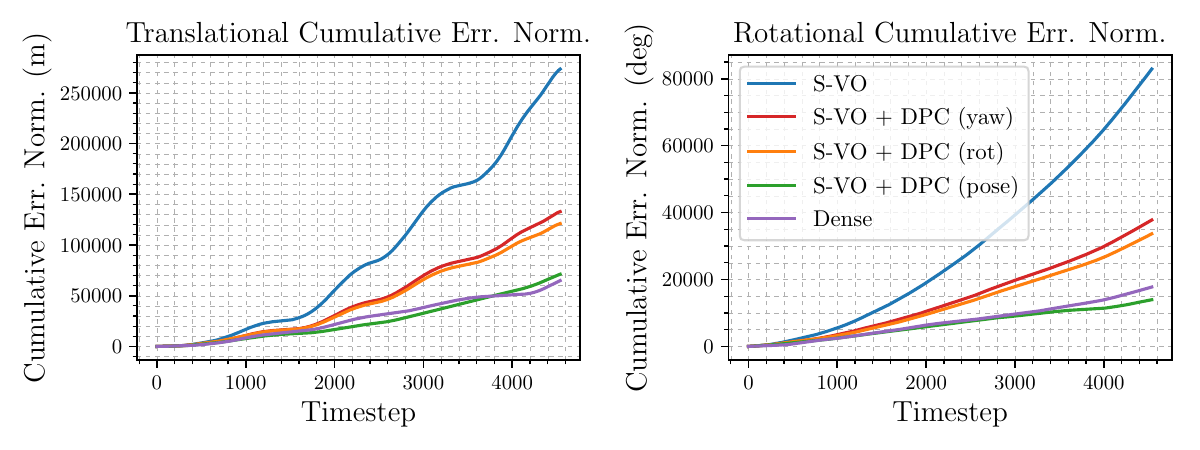}
	   \label{00-pose-cum}
	}
	\subfloat[Sequence \texttt{00}: Segment Errors.]{
		\includegraphics[width=0.5\linewidth]{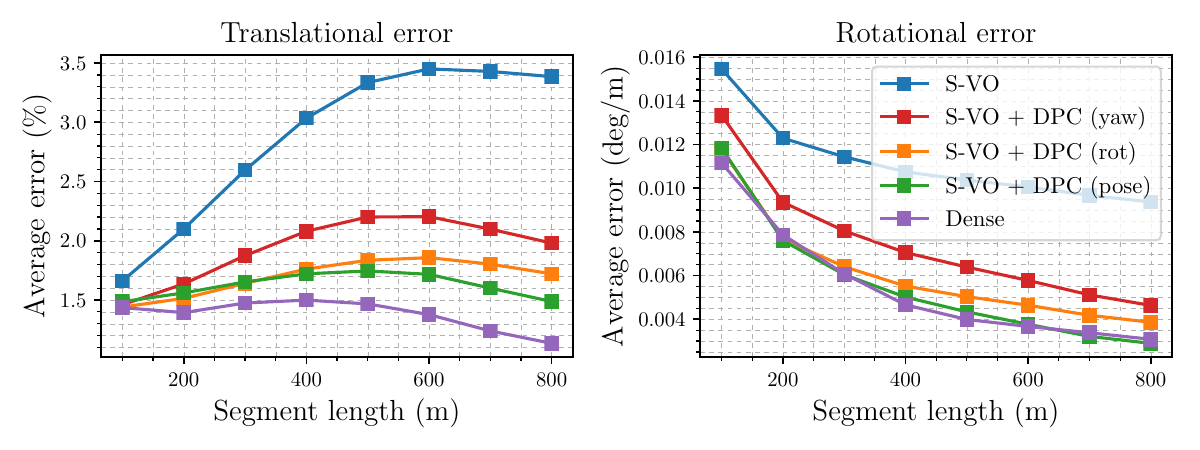}
		\label{00-pose-segs}
	} \\
	\subfloat[Sequence \texttt{02}: Cumulative Errors.]{
		\includegraphics[width=0.5\linewidth]{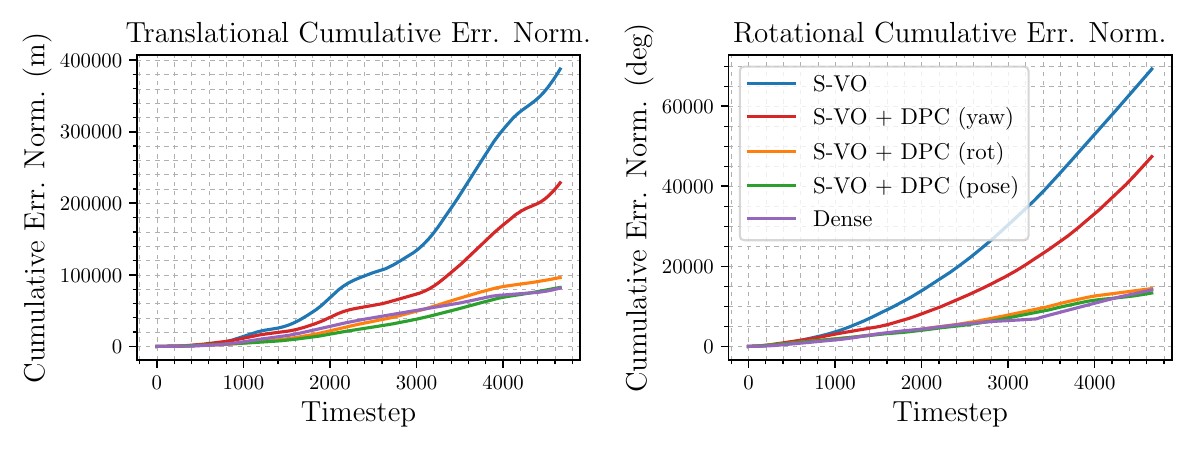}
		\label{02-pose-cum}
	 }
	 \subfloat[Sequence \texttt{02}: Segment Errors.]{
		 \includegraphics[width=0.5\linewidth]{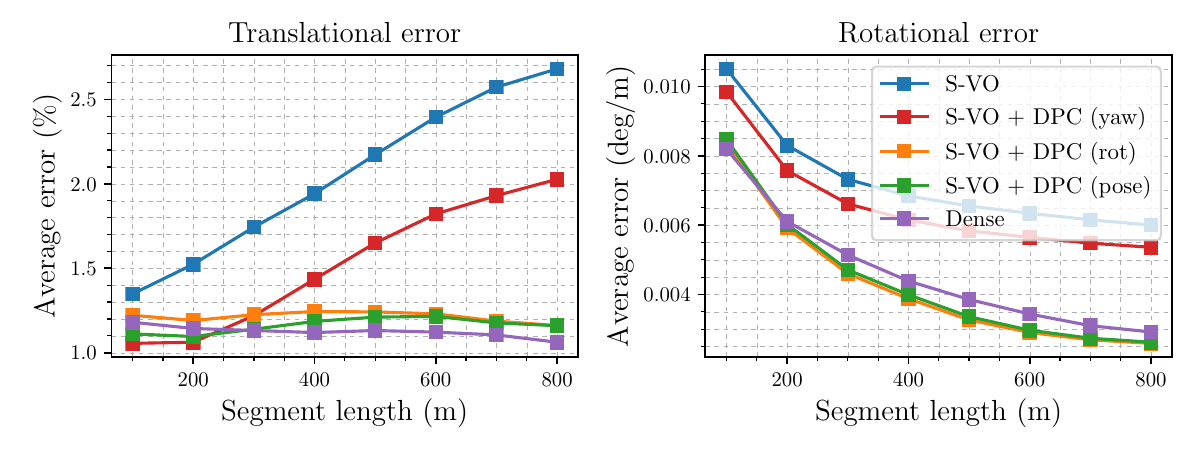}
		 \label{02-pose-segs}
	} \\
	\subfloat[Sequence \texttt{05}: Cumulative Errors.]{
		\includegraphics[width=0.5\linewidth]{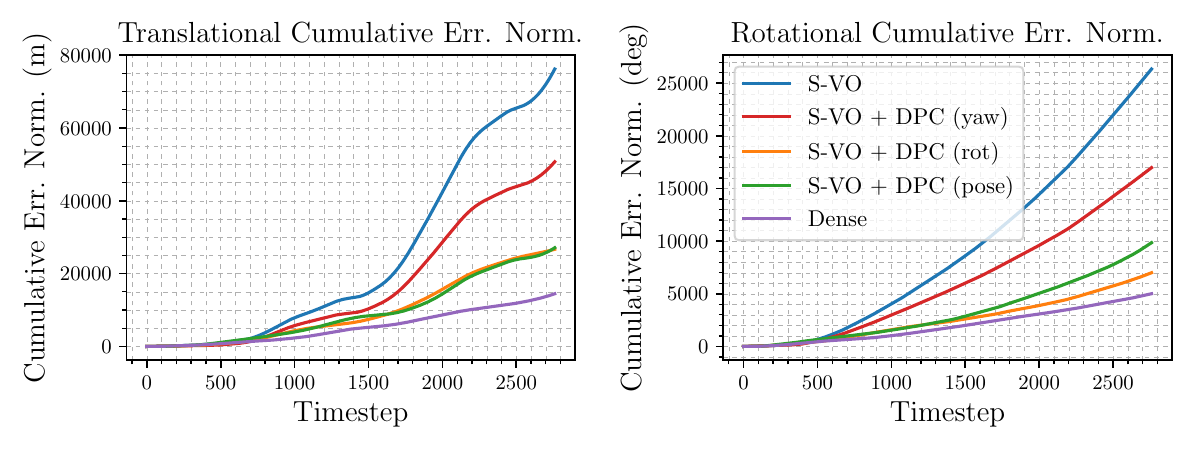}
		\label{05-pose-cum}
	}
	 \subfloat[Sequence \texttt{05}: Segment Errors.]{
		 \includegraphics[width=0.5\linewidth]{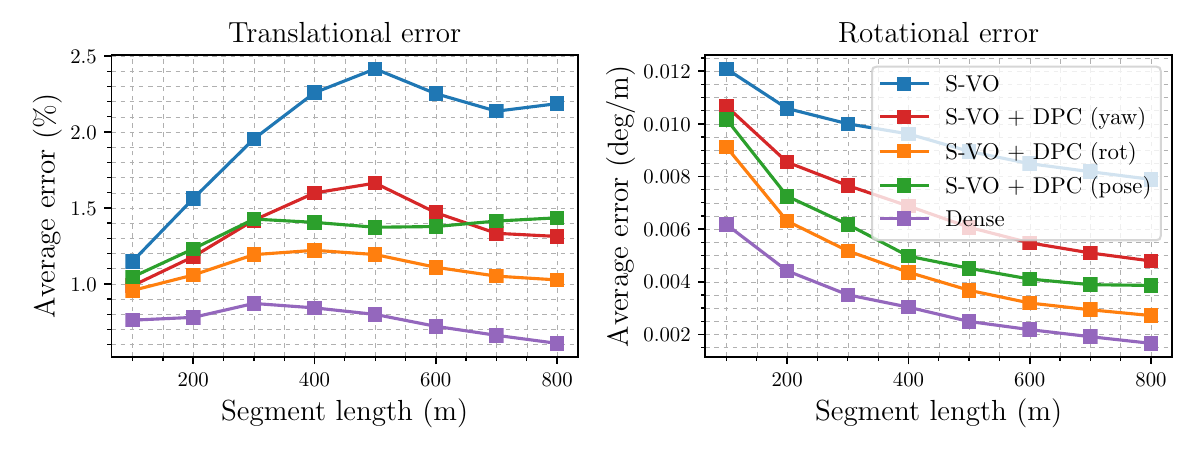}
		 \label{05-pose-segs}
	}
   \caption{c-ATE and mean segment errors for S-VO with and without DPC-Net.}
  \label{fig:cum-seg-errs} 
\end{figure*}

\begin{figure*}[h!]
	\vspace{-0.5cm}
	\centering
  	\subfloat[Sequence \texttt{00}.]{
	   \includegraphics[width=0.33\linewidth]{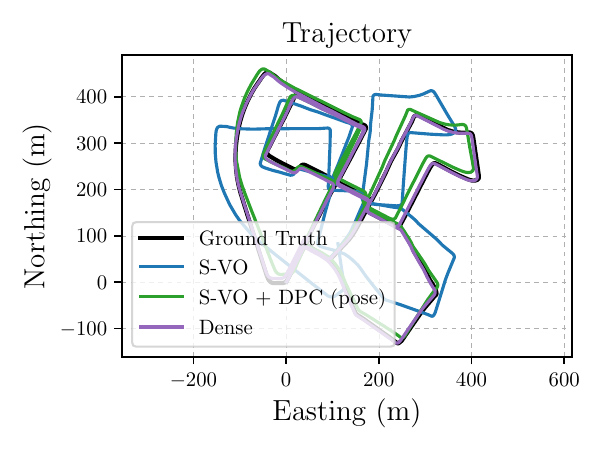}
	   \label{00-top-down}
	}
	\subfloat[Sequence \texttt{02}.]{
		\includegraphics[width=0.33\linewidth]{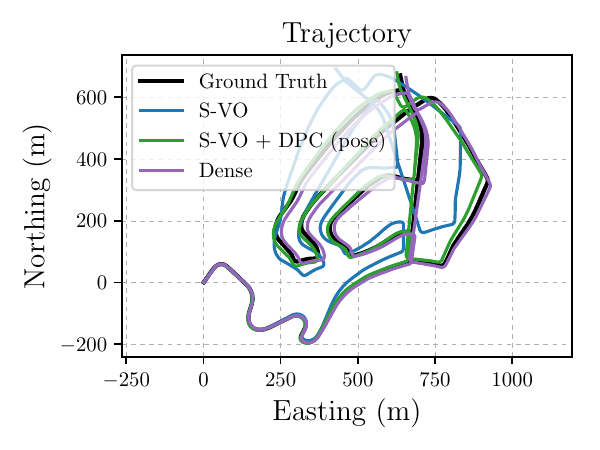}
		\label{02-top-down}
	} 
	\subfloat[Sequence \texttt{05}.]{
		\includegraphics[width=0.33\linewidth]{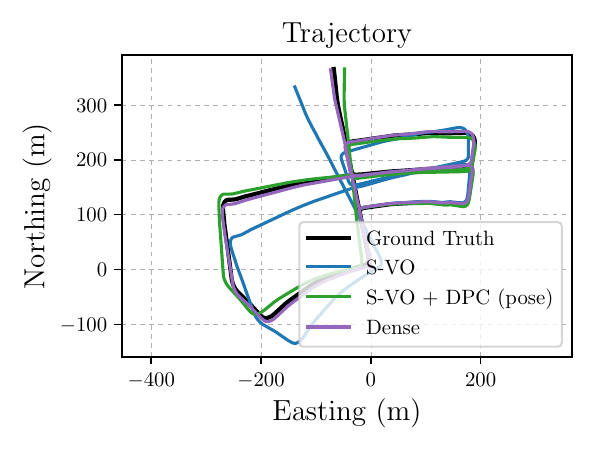}
		\label{05-pose-cum}
	 } 
   \caption{Top down projections for S-VO with and without DPC-Net.}
  \label{fig:top-downs} 
  \vspace{-0.25cm}
\end{figure*}

\subsection{Evaluation Metrics}
To evaluate the performance of DPC-Net, we use three error metrics: mean absolute trajectory error, cumulative absolute trajectory error, and mean segment error. For clarity, we describe each of these three metrics explicitly and stress the importance of carefully selecting and defining error metrics when comparing \textit{relative} localization estimates, as results can be subtly deceiving.

\subsubsection{Mean Absolute Trajectory Error (m-ATE)} The mean absolute trajectory error averages the magnitude of the rotational or translational error\footnote{For brevity, the notation $\MatLn{\cdot}$ returns rotational or translational components depending on context.} of estimated poses with respect to a ground truth trajectory defined within the same navigation frame. Concretely, $\Vector{e}_\text{m-ATE}$ is defined as
\begin{equation}
	\Vector{e}_\text{m-ATE} \definedtobe \frac{1}{N} \sum_{p=1}^{N} \Norm{\MatLn{\Estimate{\Transform}^{-1}_{p,0} \Transform_{p,0}}}.
\end{equation}
Although widely used, m-ATE can be deceiving because a single poor relative transform can significantly affect the final statistic.

\subsubsection{Cumulative Absolute Trajectory Error (c-ATE)}
Cumulative absolute trajectory error sums rotational or translational $\Vector{e}_\text{m-ATE}$ up to a given point in a trajectory. It is defined as
\begin{equation}
	\Vector{e}_\text{c-ATE} (q)  \definedtobe \sum_{p=1}^{q} \Norm{\MatLn{\Estimate{\Transform}^{-1}_{p,0} \Transform_{p,0}}}.
\end{equation}
c-ATE can show clearer trends than m-ATE (because it is less affected by fortunate trajectory overlaps), but it still suffers from the same susceptibility to poor (but isolated) relative transforms.

\subsubsection{Segment Error} Our final metric, segment error, averages the end-point error for all the possible segments of a given length within a trajectory, and then normalizes by the segment length. Since it considers multiple starting points within a trajectory, segment error is much less sensitive to isolated degradations. Concretely, $\Vector{e}_\text{seg}(s)$ is defined as
\begin{equation}
	\Vector{e}_\text{seg}(s)  \definedtobe \frac{1}{s N_s} \sum_{p=1}^{N_s} \Norm{\MatLn{\Estimate{\Transform}^{-1}_{p+s_p,p} \Transform_{p+s_p,p}}},
\end{equation}
where $N_s$ and $s_p$ (the number of segments of a given length, and the number of poses in each segment, respectively) are computed based on the selected segment length $s$. In this work, we follow the KITTI benchmark and report the mean segment error norms for all $s \in {[100, 200, 300, ..., 800]}$ (m). 

\vspace{-0.5cm}
\section{Results \& Discussion}
\begin{figure}[h!]
	\centering
	\subfloat[Sequence \texttt{00} (Distorted): Segment Errors.]{
		\includegraphics[width=\linewidth]{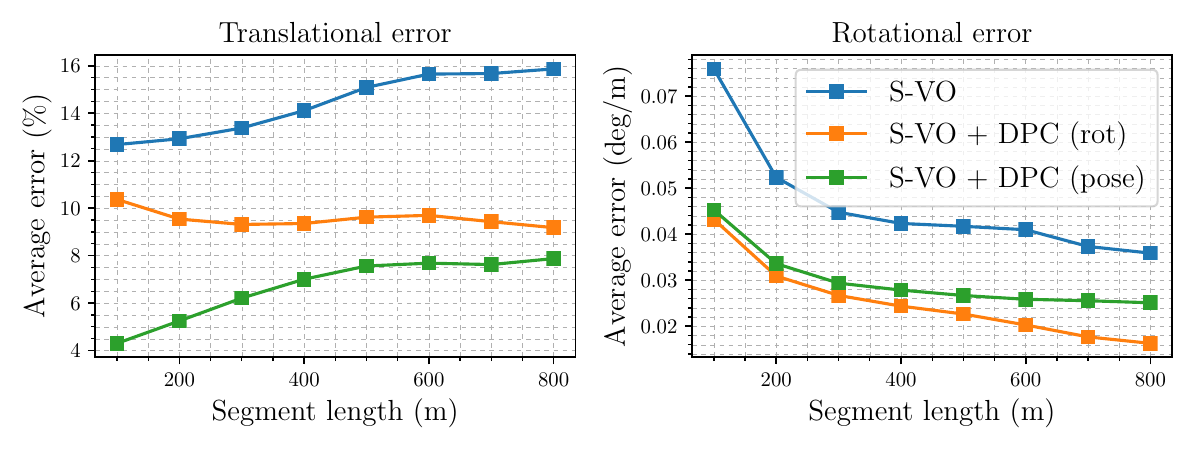}
		\label{00-pose-segs}
	} \\ 
	\subfloat[Sequence \texttt{02} (Distorted): Segment Errors.]{
		 \includegraphics[width=\linewidth]{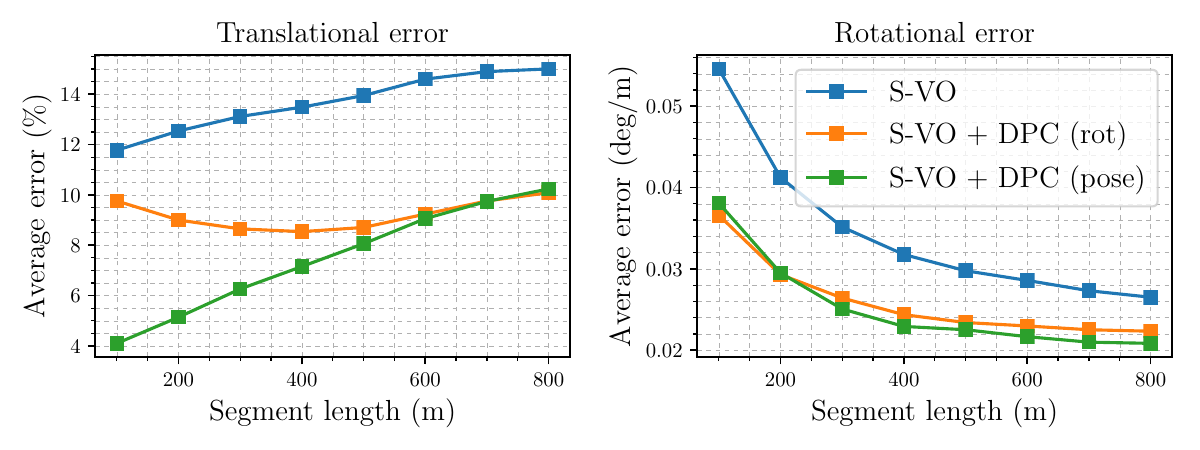}
		 \label{02-pose-segs}
	} \\
	 \subfloat[Sequence \texttt{05} (Distorted): Segment Errors.]{
		 \includegraphics[width=\linewidth]{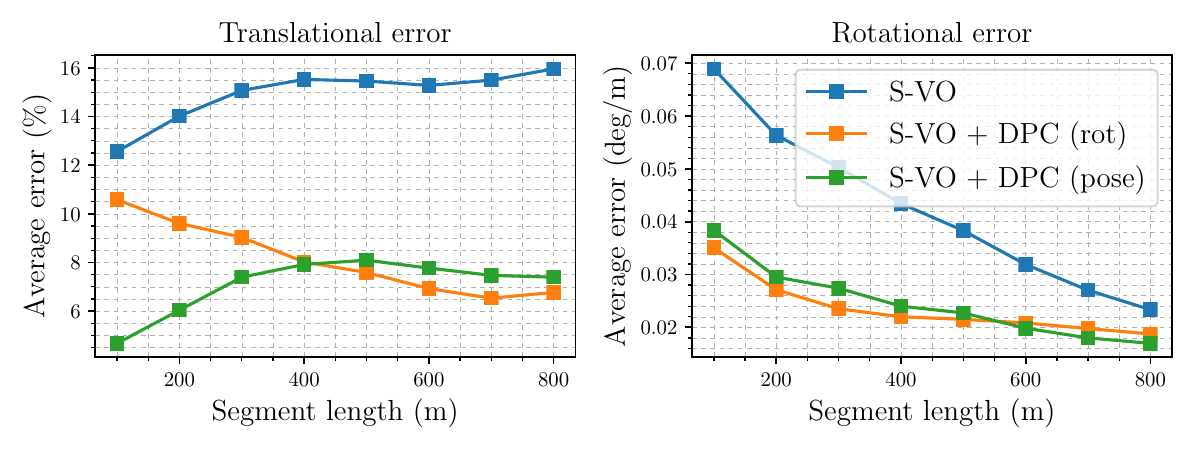}
		 \label{05-pose-segs}
	}
   \caption{c-ATE and segment errors for S-VO with radially distorted images with and without DPC-Net.}
  \label{fig:cum-seg-errs-distorted} 
\end{figure}

\subsection{Correcting Sparse Visual Odometry}
\Cref{fig:cum-seg-errs} plots c-ATE and mean segment errors for test sequences \texttt{00},  \texttt{02} and  \texttt{05} for three different DPC-Net models paired with our S-VO pipeline. \Cref{tab:nondistorted_stats} summarize the results quantitatively, while \Cref{fig:top-downs} plots the North-East projection of each trajectory.  On average, DPC-Net trained with the full $\LieGroupSE{3}$ loss reduced translational m-ATE by 72\%, rotational m-ATE by 75\%, translational mean segment errors by 40\% and rotational mean segment errors by 44\% (relative to the uncorrected estimator). Mean segment errors of the sparse estimator with DPC approached those observed from the dense estimator on sequence \texttt{00}, and outperformed the dense estimator on \texttt{02}. Sequence \texttt{05} produced two notable results: (1) although DPC-Net significantly reduced S-VO errors, the dense estimator still outperformed it in all statistics and (2) the full $\LieGroupSE{3}$ corrections performed slightly worse than their $\LieGroupSO{3}$ counterparts. We suspect the latter effect is a result of motion estimates with predominantly rotational errors which are easier to learn with an $\LieGroupSO{3}$ loss.

In general, coupling DPC-Net with a simple frame-to-frame sparse visual localizer yielded a final localization pipeline with accuracy similar to that of a dense pipeline while requiring significantly less visual data (recall that DPC-Net uses resized images).

\begin{table*}[]
    \centering
    \caption{m-ATE and Mean Segment Errors for VO results with and without DPC-Net.}
    \label{tab:nondistorted_stats}
	\begin{threeparttable}
		\begin{tabular}{lllcccc}
			\toprule
			&                            &          & \multicolumn{2}{c}{\textbf{m-ATE}}        & \multicolumn{2}{c}{\textbf{Mean Segment Errors}}   \\  \cmidrule{4-7}
\textbf{Sequence (Length)}            & \textbf{Estimator}                 &   \textbf{Corr. Type}       & Translation (m) & Rotation (deg) & Translation (\%) & Rotation (millideg / m) \\ \midrule
 \texttt{00} (3.7 km)\tnote{1} & S-VO                        & ---         & 60.22                & 18.25                & 2.88                    & 11.18                            \\
& Dense                      &  ---        & \textbf{12.41}                & \textbf{2.45}                 & \textbf{1.28}                    & \textbf{5.42}                             \\
& S-VO + DPC-Net & Pose     & 15.68                & 3.07                 & 1.62                    & 5.59                             \\
&                            & Rotation & 26.67                & 7.41                 & 1.70                    & 6.14                             \\
&                            & Yaw      & 29.27                & 8.32                 & 1.94                    & 7.47                             \\
&                            &          &                      &                      &                         &                                  \\
 \texttt{02} (5.1 km)\tnote{2} & S-VO                        &  ---        & 83.17                & 14.87                & 2.05                    & 7.25                             \\
& Dense                      &  ---        & \textbf{16.33}                & 3.19                 & 1.21                    & 4.67                             \\
& S-VO + DPC-Net & Pose     & 17.69                & \textbf{2.86}                 & \textbf{1.16}                    & 4.36                             \\
&                            & Rotation & 20.66                & 3.10                 & 1.21                    & \textbf{4.28}                             \\
&                            & Yaw      & 49.07                & 10.17                & 1.53                    & 6.56                             \\
&                            &          &                      &                      &                         &                                  \\
\texttt{05} (2.2 km)\tnote{3} & S-VO                        &   ---       & 27.59                & 9.54                 & 1.99                    & 9.47                             \\
& Dense                      &   ---       & \textbf{5.83}                 & \textbf{2.05}                 & \textbf{0.69}                    & \textbf{3.20}                             \\
& S-VO + DPC-Net & Pose     & 9.82                 & 3.57                 & 1.34                    & 5.62                             \\
&                            & Rotation & 9.67                 & 2.53                 & 1.10                    & 4.68                             \\
&                            & Yaw      & 18.37                & 6.15                 & 1.37                    & 6.90                      
\\ \bottomrule
\end{tabular}
    \begin{tablenotes}
		\item[1] Training sequences \texttt{05,06,07,08,09,10}. Validation sequence \texttt{02}. \\
		\item[2] Training sequences \texttt{00,06,07,08,09,10}. Validation sequence \texttt{05}. \\
		\item[3] Training sequences \texttt{00,02,07,08,09,10}. Validation sequence \texttt{06}. \\
		\item[4] All models trained for 30 epochs. The final model is selected based on the epoch with the lowest validation error.
    \end{tablenotes}
    \end{threeparttable}
\end{table*}

\subsection{Distorted Images}

\Cref{fig:cum-seg-errs-distorted} plots mean segment errors for the radially distorted dataset. On average, DPC-Net trained with the full $\LieGroupSE{3}$ loss reduced translational mean segment errors by 50\% and rotational mean segment errors by 35\% (relative to the uncorrected sparse estimator, see \Cref{tab:distorted_stats}). The yaw-only DPC-Net corrections did not produce consistent improvements (we suspect due to the presence of large errors in the remaining degrees of freedom as a result of the distortion procedure). Nevertheless, DPC-Net trained with $\LieGroupSE{3}$ and $\LieGroupSO{3}$ losses was able to significantly mitigate the effect of a poorly calibrated camera model. We are actively working on modifications to the network that would allow the corrected results to approach those of the undistorted case.

\section{Conclusions}
In this work, we presented DPC-Net, a novel way of fusing the power of deep, convolutional networks with classical geometric localization pipelines. Using a novel loss function based on matrix Lie groups, DPC-Net learns $\LieGroupSE{3}$ pose \textit{corrections} to improve a baseline estimator and mitigates the effect of estimator bias, environmental factors, and poor sensor calibrations. We demonstrated how DPC-Net can render a sparse stereo visual odometry pipeline as accurate as a state-of-the-art dense estimator, and significantly improve estimates computed with a poorly calibrated lens distortion model. In future work, we plan to investigate the addition of a memory state (through recurrent neural network structures), extend DPC-Net to other sensor modalities (e.g., lidar), and incorporate prediction uncertainty through the use of modern probabilistic approaches to deep learning \cite{Kendall2017-vs}.

\begin{table*}[h!]
    \centering
    \caption{m-ATE and Mean Segment Errors for VO results with and without DPC-Net for distorted images.}
    \label{tab:distorted_stats}
	\begin{threeparttable}
		\begin{tabular}{lllcccc}
			\toprule
			&                            &          & \multicolumn{2}{c}{\textbf{m-ATE}}        & \multicolumn{2}{c}{\textbf{Mean Segment Errors}}   \\  \cmidrule{4-7}
			\textbf{Sequence (Length)}            & \textbf{Estimator}                 &   \textbf{Corr. Type}         & Translation (m) & Rotation (deg) & Translation (\%) & Rotation (millideg / m) \\ \midrule
 \texttt{00-distorted} (3.7 km) & S-VO                        &   ---       & 168.27               & 37.15                & 14.52                   & 46.43                            \\
& S-VO + DPC & Pose     & 114.35               & 28.64                & \textbf{6.73}                    & 29.93                            \\
&                            & Rotation & \textbf{84.54}                & 21.90                & 9.58                    & \textbf{25.28}                            \\
&                            &          &                      &                      &                         &                                  \\
\texttt{02-distorted} (5.1 km) & S-VO                        &  ---        & 335.82               & 51.05                & 13.74                   & 34.37                            \\
& S-VO + DPC & Pose     & \textbf{196.90}               & \textbf{23.66}                & \textbf{7.49}                    & \textbf{25.20}                            \\
&                            & Rotation & 269.90               & 53.11                & 9.25                    & 25.99                            \\
&                            &          &                      &                      &                         &                                  \\
 \texttt{05-distorted} (2.2 km) & S-VO                        &  ---        & 73.44                & 12.27                & 14.99                   & 42.45                            \\
& S-VO + DPC & Pose     & \textbf{47.50}                & \textbf{10.54}                & \textbf{7.11}                    & 24.60                            \\
&                            & Rotation & 71.42                & 13.10                & 8.14                    & \textbf{23.56} \\ \bottomrule

\end{tabular}
	\end{threeparttable}
	\vspace{-1em}
\end{table*}

\printbibliography
\end{document}